\documentclass[letterpaper,times]{IONconf}
\usepackage{amsmath}
\usepackage{accents}
\newcommand{\ubar}[1]{\underaccent{\bar}{#1}}
\usepackage{natbib}
\usepackage{xfrac}

\DeclareMathOperator{\diag}{diag}

\title{Distribution of Test Statistic for Euclidean Distance Matrices}

\author{
    Dawson~Beatty
    }

\date{October 10, 2023}
\begin{document}

\maketitle
\section{Preface}
This document was originally a personal correspondence sent from Beatty to Knowles as part of a discussion on the distribution of the test statistic for a Euclidean Distance Matrix method presented in \citep{knowlesDetectionExclusionMultiple2023}. This slightly modified version has been published for the benefit of those looking to make use of EDM methods. This document has not been peer reviewed and errors may exist.

\section{Introduction}

In a recent paper about Euclidean Distance Matrices  \citep{knowlesDetectionExclusionMultiple2023}, the test statistic is defined as:
\begin{equation}
q = \frac{\lambda_4 + \lambda_5}{2 \lambda_1},
\end{equation}
where $\lambda_i$ is the $i$\textsuperscript{th} largest eigenvalue of the constructed Gram matrix $G_c$.

It is desirable to know the distribution of $q$ under nominal noise conditions so that a threshold can be set on the detection statistic. This writeup presents a derivation of this distribution and the results of a simple simulation.

\section{Test Statistic Distribution}

Given the 3D positions $\vec{r}$ of $m$ satellites combined in a matrix $X_{n \times m}$, the Gram matrix $G$ can be constructed as $G = X^T X$. From this, the Euclidean Distance Matrix (EDM) $D$ can be found as
\begin{equation}
D = \ubar{1} \diag(G)^T - 2 G + \diag(G) \ubar{1}^T.
\end{equation}
At the time instant under consideration, the EDM for the satellite positions can be computed using almanac data. The user additionally has access to $m$ pseudorange measurements of the form
\begin{equation}
\rho_i = d_i + b + v_i, \hspace{1cm} \text{for $i$ between $1$ and $m$}
\end{equation}
where $d_i$ is the distance between the user and the satellite, $b$ is the range-equivalent receiver clock bias, and $v_i$ are independent and identically distributed normal random variables with zero mean and standard deviation $\sigma_v$. For now, assume that the clock bias is known and has been subtracted out of the measurements, so the pseudoranges are just true distance plus noise.

After the pseudorange measurements have been obtained, the EDM can be constructed as
\begin{equation}
D_c = 
\begin{bmatrix}
0 & 
\begin{bmatrix}
\rho_1^2 & \rho_2^2 & \cdots & \rho_m^2
\end{bmatrix} \\
\begin{bmatrix}
\rho_1^2\\ \rho_2^2 \\ \vdots \\ \rho_m^2
\end{bmatrix} & 
[D]
\end{bmatrix}.
\end{equation}
This can be converted to the Gram matrix using
\begin{equation}
G_c = -\frac{1}{2} J D_c J \hspace{0.5cm} \text{where} \hspace{0.5cm}
J = I - \frac{1}{m} \ubar{1}\ubar{1}^T.
\end{equation}
After this matrix is constructed, the eigenvalues $\{ \lambda_i \}_{i=1}^{m+1}$ are computed and the test statistic is generated. If no faults or measurement noise are present, then there will be three non-zero eigenvalues ($\lambda_1, \lambda_2, \lambda_3)$ with a large value, with all of the rest being zero. When the distances in the distance matrix are not all consistent, resulting from noise or a fault, the resulting Gram matrix will have two additional non-zero eigenvalues, $\lambda_4$ and $\lambda_5$. The effect of faults will not be considered in this paper. The first step of the process for determining the distribution of the eigenvalues under nominal noise conditions is determining the perturbation to the eigenvalues introduced by noise.

\subsection{Eigenvalue perturbation}

Call the nominal Gram matrix constructed above $G_{c, \text{ nom}}$. Each of the pseudorange measurements that went into $D_c$ has some noise. For purposes of analysis, consider a version of the EDM with the noise subtracted out:
\begin{equation}
D_c = 
\begin{bmatrix}
0 & 
\begin{bmatrix}
(\rho_1 - v_1)^2 & (\rho_2 - v_2)^2 & \cdots & (\rho_m - v_m)^2
\end{bmatrix} \\
\begin{bmatrix}
(\rho_1 - v_1)^2\\ (\rho_2 - v_2)^2 \\ \vdots \\ (\rho_m - v_m)^2
\end{bmatrix} & 
[D]
\end{bmatrix}.
\end{equation}
This can be used to construct the associated Gram matrix, called $G_{c, \text{ noiseless}}$. The effect of noise on the Gram matrix can be approximated using a Taylor series:
\begin{align}
G_{c, \text{ nom}} 
&\approx G_{c, \text{ noiseless}} + \sum\limits_{i=1}^m \left.\frac{\partial G_{c, \text{ noiseless}}}{\partial v_i} \right|_{v_i = 0} v_i \\
&\approx G_{c, \text{ noiseless}} + \sum\limits_{i=1}^m \delta G_{c, i} v_i 
\end{align}
For GNSS purposes, the noise tends to be smaller than the distances by many orders of magnitude, so this approximation is accurate. Consider the eigenvalue problem 
\begin{equation}
G \ubar{z} = \lambda \ubar{z}.
\end{equation}
The goal of this section is to determine the effect of a perturbation to $G$ on a given eigenvalue pair $\lambda_0$ and $\ubar{z}_0$. Consider a perturbation $\varepsilon V$ acting on $G$. The new eigenproblem takes the form
\begin{equation}
(G + \varepsilon V) \ubar{z} = \lambda \ubar{z}.
\end{equation}
\cite{hinchPerturbationMethods1991} shows that for unique non-zero roots, the first-order effect of the perturbation on the eigenvalue will take the form
\begin{equation}
\lambda \approx \lambda_0 + \frac{\ubar{z}_0^T V \ubar{z}_0}{\ubar{z}_0^T \ubar{z}_0} \varepsilon.
\end{equation}
In the case of EDMs applied to GNSS, $(\lambda_0, \ubar{z}_0)$ are an eigenvalue/eigenvector pair computed from $G_{c, \text{ nom}}$ and the small perturbation will be the noise $v$. For each of the first three nominally non-zero eigenvalues, the first-order perturbations will be
\begin{equation}
\lambda_{i} \approx \lambda_{i, \text{ nominal}} + \sum\limits_{j=1}^m \frac{\ubar{z}_i^T \delta G_{c, j} \ubar{z}_i}{\ubar{z}_i^T \ubar{z}_i} v_j.
\end{equation}
The expected value of the noise $v$ is zero, so it is simple to show that the expected value of the perturbed eigenvalue is the same as the nominal. The variance can be computed similarly:
\begin{equation}
E[(\lambda_{i} - \lambda_{i, \text{ nominal}})^2] = 
\sum\limits_{i=1}^m 
\left( \frac{\ubar{z}_i^T \delta G_{c, j} \ubar{z}_i}{\ubar{z}_i^T \ubar{z}_i} \sigma_v \right)^2
\end{equation}
The noise terms are assumed to be independent, so the cross terms in the expected value all go to zero, and the variance of the perturbed eigenvalue is the weighted sum of the squares of normal random variables. This means that each eigenvalue is distributed as well under nominal noise conditions.

But wait! One of the caveats of the eigenproblem perturbation method was unique non-zero eigenvalues! The eigenvalues $\lambda_4$ and $\lambda_5$ are normally zero! Using the current assumptions, the method above won't work. The problem is that the eigenvectors associated with $\lambda_4$ and $\lambda_5$ are dependent on the exact noise values, and it is difficult or impossible to predict what the eigenvectors will be. 

So the problem is that $\lambda_4$ and $\lambda_5$ are zero without noise and without the clock bias. If one does not subtract out the clock bias, the two values will be unique and non-zero. In Knowles's paper it is assumed that the clock bias is known, so it can just be left in. If one does not subtract out the clock bias, the two values will be unique and non-zero. In Knowles's paper it is assumed that the clock bias is known, so it can just be left in.  

This works because including the clock bias sets the eigenvector, and so the perturbation method works. The noise will still modify the eigenvector a little though. For this reason, it might sometimes make sense to artificially inflate the clock bias term to ensure that the noise won't significantly modify the eigenvectors.

\subsection{Test Statistic}

The previous subsection showed that, under nominal noise conditions, the eigenvalues will be normally distributed. The subsection also showed how to determine the variance of each of the eigenvalues.

The eigenvalues appear to be statistically independent of each other from plotting a scatter plot with a large number of simulations.

The test statistic is defined as
\begin{equation}
q = \frac{\lambda_4 + \lambda_5}{2 \lambda_1}.
\end{equation}
The distribution of the numerator is easy. Gaussian random variable (RV) plus another Gaussian RV is Gaussian as well. The ratio of two Gaussian RVs is more challenging. The resulting RV does not have finite moments and is heavy tailed. This paper \citep{diaz-francesExistenceNormalApproximation2013} that says that given two random variables $X$ and $Y$, the distribution of $Z = \sfrac{X}{Y}$ is Gaussian under some conditions. Further testing is needed, but a Gaussian with their method seems to fit the data well. If $X \sim \mathcal{N}(\mu_x, \sigma_x^2)$ and $Y \sim \mathcal{N}(\mu_y, \sigma_y^2)$, then
\begin{equation}
Z \sim \mathcal{N} \left( \frac{\mu_x}{\mu_y}, \frac{\mu_x^2}{\mu_y^2} \left( \frac{\sigma_x^2}{\mu_x^2} + \frac{\sigma_y^2}{\mu_y^2} \right) \right).
\end{equation}

\section{Simulated Results}

A python script \texttt{simulation\_3D.py} will be included with this file. It considers a point solution with $m=12$ satellites with realistic geometry. 

\begin{figure}[h]
\centering
\includegraphics[width=8cm]{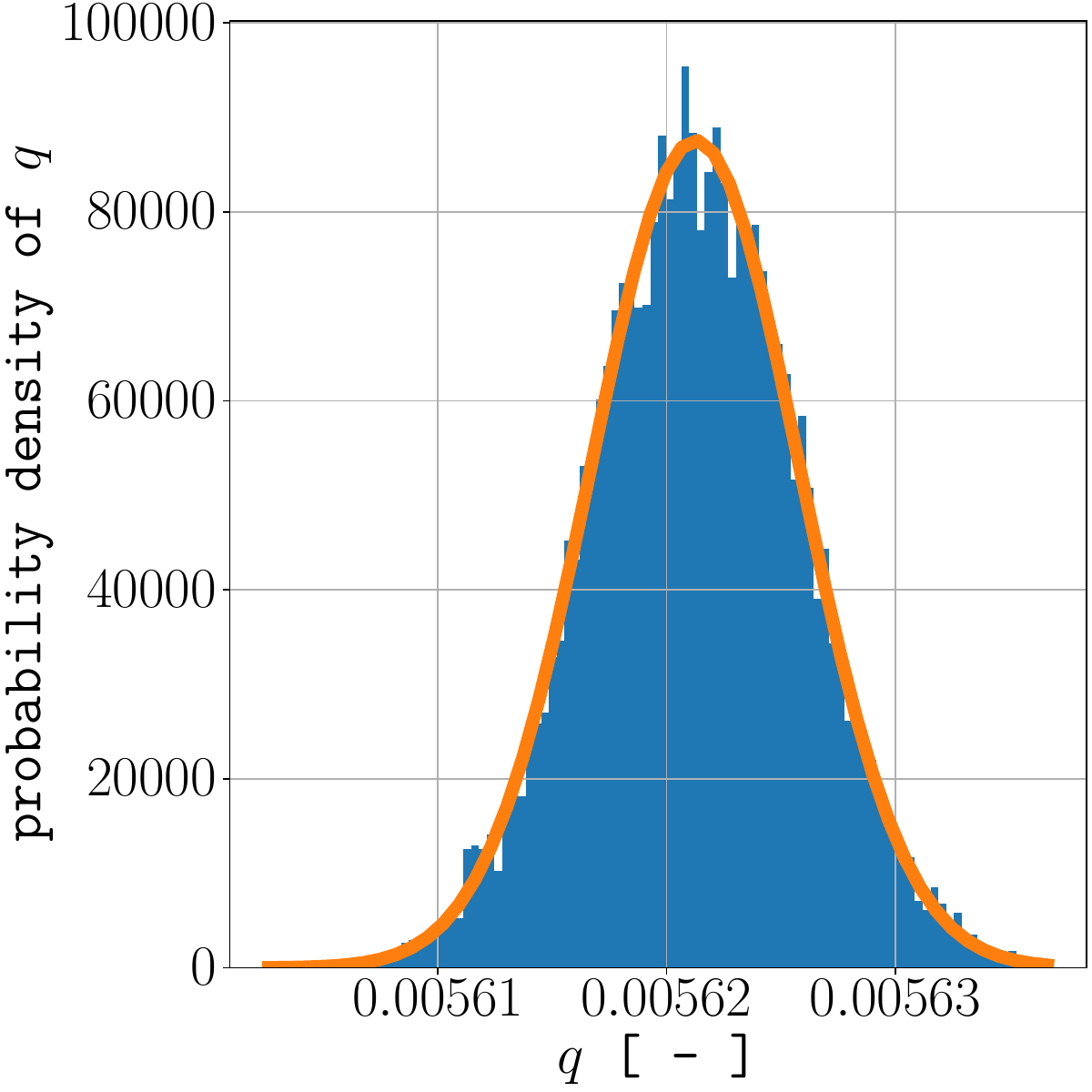}
\caption{Comparison of empirical distribution of test statistic (blue bins) against theoretical normal distribution (orange line)}
\label{fig:q}
\end{figure}

The simulation was run $10,\!000$ times with the same geometry but different noise values. The histogram in Figure \ref{fig:q} shows the empirical distribution of the test statistic from each of these runs. The figure also shows an orange line, which is the Gaussian approximation to the ratio distribution, as discussed in the previous section. 

\section{Conclusion}

The results presented in this paper should be sufficient to determine the nominal distribution of the test statistic at a given geometry. This will allow the algorithm designer to determine a test statistic for fault detection.

This investigation is only a proof of concept, and further validation would be needed. The ratio distribution is well-approximated as a Gaussian only under specific conditions, and further research would be needed to determine whether this method would always meet those conditions in GNSS scenarios. 

The author strongly suspects that this method would not perform as well when the measurement noise $v$ is larger compared to the pseudorange measurements. In the case of GNSS, it is always the case that $\rho \gg v$ and so the exact distribution of the noise does not significantly perturb the dominant three eigenvectors. That may not be the case in scenarios where the user and transmitters are much closer to each other.

Similarly, determination of the distribution of the fourth and fifth eigenvalues requires inclusion of the receiver clock bias term in the pseudorange (or introducing fake bias to stabilize the eigenvectors). Too small of a bias will not stabilize the eigenvectors properly, while too large of a bias may introduce errors in some other way (although the author is not sure how).

One of the advantages of the EDM method is the execution time \citep{knowlesEuclideanDistanceMatrixBased2023}, but this advantage may be reduced when evaluating the distribution of the test statistic. The form of $\delta G_c$ can be determined offline, but it will need to be evaluated at the current geometry. This is still a fast operation, but the EDM-based method may lose its advantage over residual-based RAIM in terms of execution time.

The method presented here assumes that the fault-free pseudorange measurements are known. Future work would include discussion of determination of the test statistic distribution in the presence of faults. 

\bibliographystyle{apalike}
\bibliography{edm.bib}

\end{document}